\documentclass[letterpaper, 10 pt, conference]{ieeeconf}  

\IEEEoverridecommandlockouts                              

\usepackage{multicol}
\usepackage[bookmarks=true]{hyperref}
\usepackage[dvipsnames]{xcolor}
\usepackage{graphicx}
\usepackage{subcaption}
\usepackage{comment}
\usepackage{amsmath}
\usepackage{amssymb}
\usepackage{mathtools}
\usepackage{xspace}
\usepackage{algorithm}
\usepackage{algpseudocode} 
\usepackage{bm} 
\usepackage{booktabs}
\usepackage[%
  backend=biber,
  url=true,
  style=ieee, 
  sorting=none, 
  maxnames=4,
  minnames=3,
  maxbibnames=99,
  giveninits,
  uniquename=init]{biblatex} 
\usepackage{censor}

\newif\ifarxiv
\arxivfalse

\renewcommand{\baselinestretch}{0.95}

\title{\LARGE \bf
ADMM-based Continuous Trajectory Optimization \\ in Graphs of Convex Sets
}

\author{Lukas Pries$^{1}$, Jon Arrizabalaga$^{2}$, Zachary Manchester$^{2}$ and Markus Ryll$^{1}$
\thanks{$^{1}$Lukas Pries and Markus Ryll are with the Autonomous Aerial Systems Lab, Dep. of Aerospace and Geodesy,
        TU Munich, Germany,
        {\tt\small lukas.pries@tum.de, markus.ryll@tum.de}}%
\thanks{$^{2}$Jon Arrizabalaga and Zachary Manchester are with the Department of Aeronautics and Astronautics, Massachusetts Institute of Technology, USA
        {\tt\small jonarri@mit.edu, zacm@mit.edu }}
}

\bibliography{references}

\begin{document}

\newcommand{\nextpage}{%
  \newpage
  \hspace{1cm}%
  \newpage
}

\newcommand{\acronym}{ACTOR\xspace}

\newcommand{\vv}{q}

\setlength{\textfloatsep}{6pt}
\setlength{\floatsep}{6pt}
\setlength{\intextsep}{6pt}



\maketitle
\thispagestyle{empty}
\pagestyle{empty}

\begin{abstract}

This paper presents a numerical solver for computing continuous trajectories in non-convex environments. Our approach relies on a customized implementation of the Alternating Direction Method of Multipliers (ADMM) built upon two key components: First, we parameterize trajectories as polynomials, allowing the primal update to be computed in closed form as a minimum-control-effort problem. Second, we introduce the concept of a spatio-temporal allocation graph based on a mixed-integer formulation and pose the slack update as a shortest-path search. The combination of these ingredients results in a solver with several distinct advantages over the state of the art. By jointly optimizing over both discrete and continuous domains, our method accesses a larger search space than existing decoupled approaches, enabling the discovery of superior trajectories. Additionally, the solver's structural robustness ensures reliable convergence from naive initializations, removing the bottleneck of complex warm starting in non-convex environments.

\end{abstract}

\section{INTRODUCTION}

Simultaneous discrete and continuous search is fundamental to various scientific and engineering domains, including task and motion planning, hybrid system control, and constrained decision-making. In these settings, discrete choices—such as selecting topological routes or contact modes—are intrinsically coupled with continuous variables like system states and control inputs. Given this interdependence, algorithms capable of efficiently addressing this problem class could have a profound impact across a vast portfolio of autonomous applications.

Historically, the joint search over discrete and continuous spaces was considered prohibitively difficult, leading to a traditional decoupling of planning (high-level logic) and control (low-level execution). This separation allows for specialized numerical techniques: sampling- or search-based methods for the discrete planning stage and gradient-based optimization for the continuous control stage. However, this ``numerical convenience'' comes at a significant cost. By partitioning the search space, the optimization becomes blind to the global landscape, often resulting in holistically suboptimal performance.

\begin{figure}[t]
    \centering
    
    \begin{subfigure}{0.27\textwidth}
        \centering
        \includegraphics[width=\linewidth]{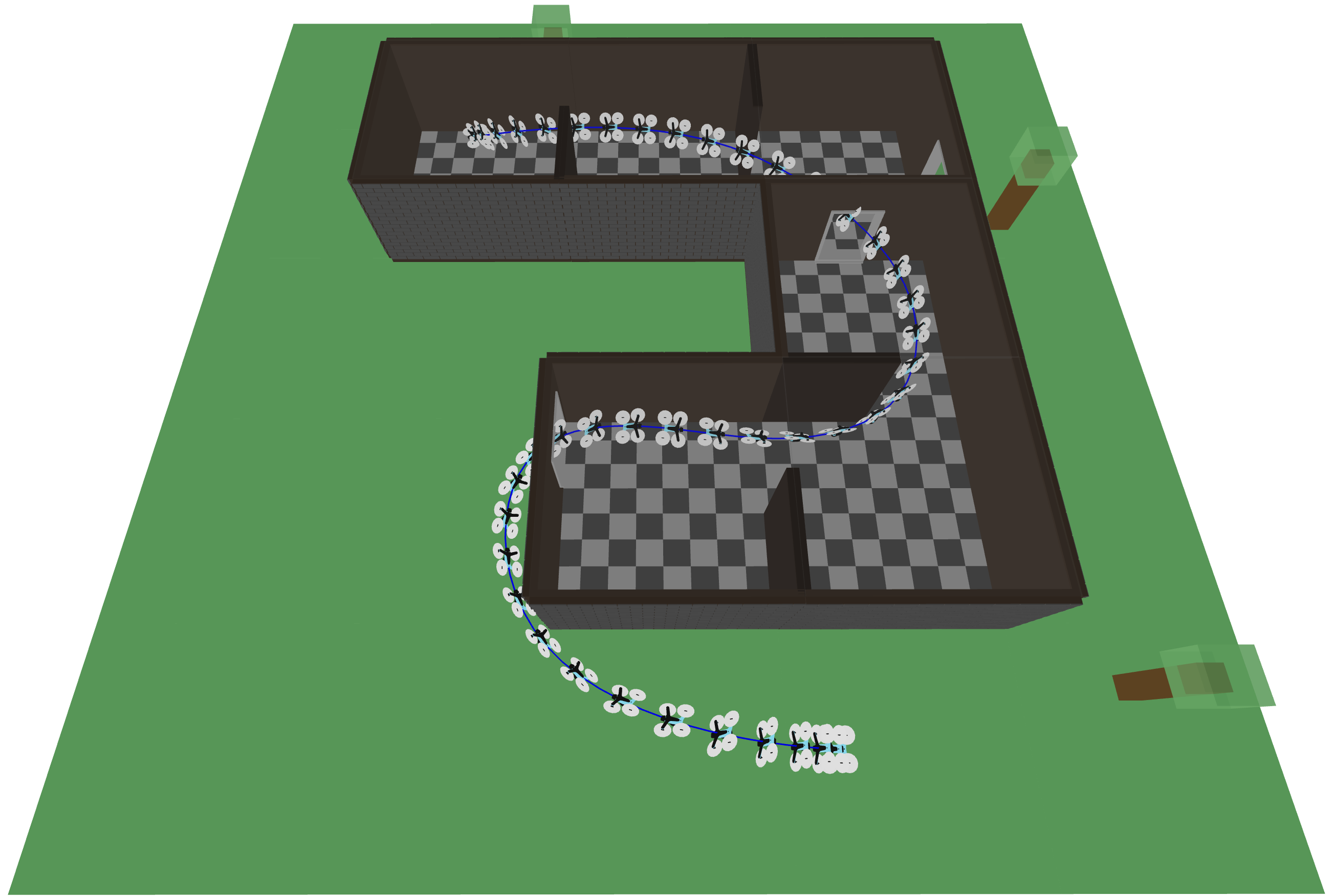}
    \end{subfigure}
    \hfill
    \begin{subfigure}{0.19\textwidth}
        \centering
        \includegraphics[width=\linewidth]{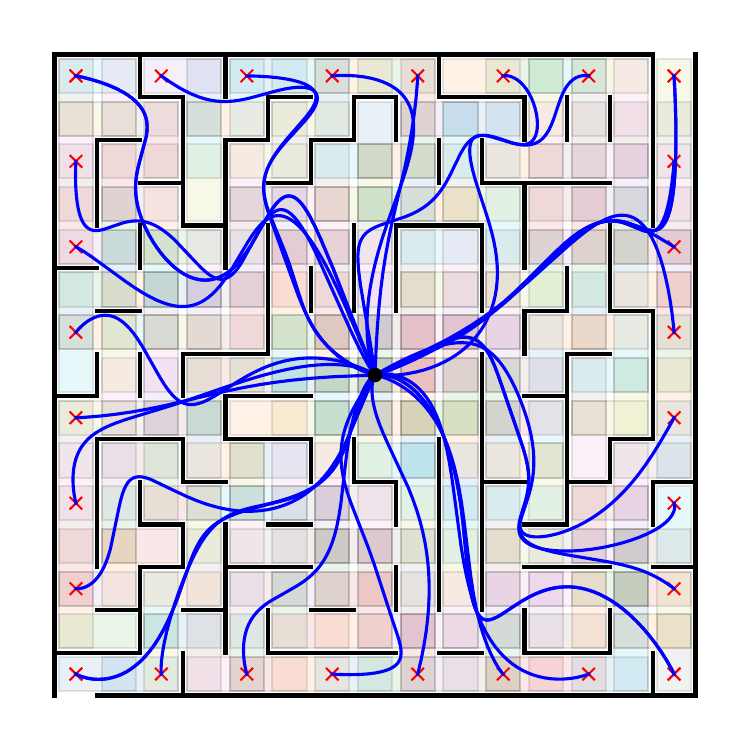}
    \end{subfigure}
    \caption{Example trajectories generated by ACTOR: a quadrotor minimum-snap trajectory (left) and high-order continuous trajectories through a non-convex maze (right).}
    \label{fig:quad}
\end{figure}
To address these challenges, this paper introduces a novel numerical solver that leverages a customized Alternating Direction Method of Multipliers (ADMM) for continuous trajectory optimization over a union of convex sets. This allows us to decompose the joint optimization into primal, slack, and dual updates, each tailored to exploit the specific mathematical structure of the problem.

Our solver, which we denote as \textbf{ACTOR} (\textbf{A}DMM-based \textbf{C}ontinuous \textbf{T}rajectory \textbf{O}ptimize\textbf{R}), is built upon the following key contributions:

\begin{enumerate}
    \item We introduce the concept of a \textit{spatio-temporal allocation graph} based on a mixed-integer formulation of the non-convex safety constraints. The graph encodes all feasible convex realizations of the problem and provides the structure required for projection onto the non-convex feasible set, thereby enabling an ADMM-based solver.

    \item By formulating the primal update as a \textit{quadratic minimum-control-effort problem} over polynomial segments, we derive a closed-form solution. The factorization can be performed offline, significantly reducing the per-iteration computational overhead.
    
    \item We frame the slack update as a \textit{shortest-path problem} over the weighted allocation graph. Since the graph is directed and acyclic, this search procedure reduces to a simple dynamic programming iteration with linear computational complexity.
\end{enumerate}

The remainder of this paper is structured as follows: Section~\ref{sec:problem_statement} formally defines the problem, while Section~\ref{sec:preliminaries} introduces the necessary mathematical background. We detail our proposed method in Section~\ref{sec:method} and evaluate its performance through various benchmarks in Section~\ref{sec:experiments}. Finally, Section~\ref{sec:conclusion} provides concluding remarks and discusses directions for future work.

\section{RELATED WORK}

\subsection{Nonlinear Programming}
A general approach to collision-free trajectory optimization in non-convex environments is to formulate collision avoidance constraints through direct nonlinear constraints between the trajectory and the environment.
These formulations result in nonlinear programs which can be solved using general nonlinear solvers based on Newton-type methods such as interior-point (e.g. IPOPT \cite{biegler2009large}) and sequential convex programming (SNOPT \cite{gill2005snopt}). 
However, formulating \textit{smooth nonlinear constraints} with respect to general environment representations is non-trivial.
In robotic motion planning, Ratliff et al. \cite{ratliff2009chomp} first introduced Euclidean signed-distance fields (ESDF) to obtain gradient information w.r.t. obstacles.
The SQP-based solver TrajOpt by Schulman et al. \cite{schulman2013finding} considers signed-distance (SD) functions to convex obstacles and propose a geometric linearization procedure based on separating hyperplanes. Following work by Zhang et al. \cite{zhang2020optimization} reformulates the generally non-differentiable SD constraints into an exact smooth formulation by exploiting the strong duality for convex sets.
A key limitation of these methods is their susceptibility to \textit{infeasible local minima} affecting their reliability and completeness \cite{schulman2013finding, ratliff2009chomp}.
\subsection{Mixed-Integer Programming}
Mixed-integer programming offers a systematic way to encode exact obstacle avoidance constraints in smooth trajectory optimization.
By modeling the non-convex set of obstacle-free states as the union of finitely many convex regions, one can perform a mixed-integer convex optimization in which the integer variables correspond to the assignment of trajectory segments to convex regions. 
This formulation is desirable for several reasons. First, the arbitrary connection of sets admits exploration of many different possible paths in a non-convex environment \cite{deits2015efficient}.
Second, the flexible assignment of different numbers of (fixed-duration) segments to individual sets ensures spatio-temporal flexibility while avoiding nonlinear optimization over time as proposed by FASTER \cite{tordesillas2021faster}.
However, the \textit{combinatorial nature} of the mixed-integer problem makes it NP-hard and any algorithm that finds the global solution suffers from non-polynomial worst-case runtime \cite{takapoui2020simple}. This prevents the application of MIQP beyond small problem instances. The seminal work by Marcucci et al. \cite{marcucci2023motion} presents an empirically-tight convex relaxations of this problem based on Semi-Definite Programming. However, the method is restricted to objectives and bounds involving derivatives up to velocity and does not accommodate \textit{higher-order smoothness} objectives or constraints.

\subsection{Decoupled Methods} 
Another widely adopted approach is to decompose the problem into a series of simpler subproblems.
In the first stage, these approaches usually leverage shortest-path heuristics \cite{marcucci2024fast, ren2025safety} to heuristically solve the discrete and non-convex part of the planning problem and, only at a later stage, use continuous optimization to shape the trajectory in the selected corridor. Many works on autonomous navigation employ graph-search methods (e.g. A* \cite{ren2025safety} or RRT* \cite{richter2016polynomial}) to find shortest paths on occupancy grids and subsequently construct a safe corridor. If there already exists a decomposition of the environment, the problem may be posed as a shortest-path problem in a Graph of Convex Sets (GCS) \cite{marcucci2024fast}. Von Wrangel and Tedrake \cite{von2024using} employ this method with convex surrogate cost for safe corridor selection and initialization of the non-convex setting involving constraints and cost on higher-order derivatives.
In the subsequent stage, a continuous spatio-temporal trajectory is optimized inside a sequence of convex sets (corridor). The nonlinear relationship with respect to time still renders this a nonlinear programming (NLP) problem \cite{von2024using}. 
Tordesillas et al. avoid optimizing over time and ensure temporal flexibility using a MIQP formulation \cite{tordesillas2021faster}.
The most efficient approach is to reformulate the NLP into an unconstrained problem by relaxing constraints into soft penalties \cite{wang2022geometrically}. However, these methods rely on cumbersome tuning for different scenarios and cannot guarantee constraint satisfaction.

In our work, we adopt the general mixed-integer formulation which allows for flexible path and time allocation. Building upon this formulation, we
present a novel gradient-based method based on ADMM \cite{takapoui2020simple} which is reliable and efficient in finding locally optimal solutions to this problem.

\section{PROBLEM STATEMENT} \label{sec:problem_statement}
In this section we introduce the problem formulation and the underlying assumptions that we make in order to solve the problem. We consider the following trajectory generation problem:
\begin{subequations} \label{eq:problem_statement_0}
\begin{align}
    \min_{\vv}& & &J(\vv) & \text{(efficiency)}\\
    \text{subject to}& & & \vv(0) = \vv_0, \: \vv(T) = \vv_T & \text{(boundary)} \label{eq:1boundary} \\
    &&& \vv \in \mathcal{D} & \text{(feasibility)} \label{eq:1feasibility} \\
    &&& \vv \in \mathcal{S} & \text{(safety)} \label{eq:1safety},
\end{align}
\end{subequations}
where $\vv : [0,T] \rightarrow \mathbb{R}^d$ represents a continuous, sufficiently differentiable trajectory.
Constraint~\eqref{eq:1safety} enforces collision avoidance by requiring the trajectory $\vv$ to remain within the obstacle-free space \( \mathcal{S} \subset \mathbb{R}^d \), which is generally non-convex. Dynamic feasibility is imposed through constraint~\eqref{eq:1feasibility}, where \( \mathcal{D} \) denotes the set of dynamically admissible trajectories.
Finally, the constraints in \ref{eq:1boundary} enforce the boundary conditions for the start and goal positions.
Due to the continuous-time formulation and the presence of nonlinear and non-convex constraints, solving problem~\eqref{eq:problem_statement_0} directly is computationally intractable. We therefore introduce structural assumptions that render the problem numerically well posed and amenable to efficient optimization.

\paragraph{Objective -- Minimum Control Effort}
We consider smooth trajectories that minimize the control effort required for the robot to follow these trajectories. The objective is therefore a weighted sum of integral quadratic penalties on the derivatives of the trajectory up to $n$-th order: 
\begin{align}\label{eq:cost}
    J(q) = \sum_{i=1}^{n} \alpha_i \int_0^{T} ||\vv^{(i)}(t)||^2 dt
\end{align}

\paragraph{Dynamic Feasibility -- Continuous Trajectories}
\ifarxiv
We consider differentially flat systems, including any controllable linear systems \cite{brunovsky1970classification} and nonlinear systems that are locally linearizable by dynamic feedback \cite{fliess1995flatness}.
\else
We consider differentially flat systems, including any controllable linear system and nonlinear systems that are locally linearizable by dynamic feedback.
\fi
Any sufficiently differentiable trajectory of the flat outputs is dynamically feasible, provided their derivatives are sufficiently bounded to avoid input saturation. For nonlinear systems, linear bounds may be conservative but generally provide a practical approximation for trajectory generation.
Accordingly, we only require the trajectory $q: [0,T] \mapsto \mathbb{R}^d$ to be continuously differentiable and bounded up to sufficiently high orders $n_c$ and $n_b$: 
\begin{equation}
    \mathcal{D} := \left\{ \vv \; | \; \vv \in \mathcal{C}^{n_c}, \; \vv^{[n_b]}(t) \in \mathcal{B}, \: \forall t \in [0,T] \right\}, \label{eq:feas}
\end{equation}
 with $\vv^{[n_b]}=(\dot{\vv},\ddot{\vv},\ldots,\vv^{(n_b)})^\top$ and the convex bounded set $\mathcal{B}$.

\paragraph{Safety -- Union of Convex Sets}
Without loss of generality, we represent the non-convex search space through a union of convex sets as
\begin{align}\label{eq:safe}
    &\mathcal{S} := \left\{ \vv \; | \; \vv(t) \in \bigcup\nolimits_{k \in \mathcal{K}} \mathcal{Q}_k, \: \forall t \in [0,T] \right\}  \;,
\end{align}
where $\mathcal{K}$ is a finite index set and each $\mathcal{Q}_k$ denotes a convex subset of the free space. We only require $\mathcal{Q}_k$ to be simple in the sense that Euclidean projections onto $\mathcal{Q}_k$ can be performed efficiently (e.g., ball sets, box sets, or polyhedra in low-dimensional spaces).

In the following, we introduce a transcription of the continuous state variables $\vv$ \eqref{eq:feas} and address the safety constraint over a union of convex sets \eqref{eq:safe} through a mixed-integer formulation.

\section{METHOD}\label{sec:preliminaries}
The solver presented in this work relies on two main building blocks: (i) a piecewise-polynomial parameterization of the system dynamics and (ii) a graph-based representation of the feasible space through an allocation graph. This section introduces these components and formalizes the required definitions before presenting the solver in the next section.

\subsection{Trajectories as Piecewise Polynomials}\label{subsec:polynomials}

To solve problem~\eqref{eq:problem_statement_0} numerically, the trajectory $\vv(t)$ must be parameterized using a finite number of decision variables. 
Due to their smoothness and differentiability properties, polynomials provide a natural choice for this parameterization. Moreover, the feasibility and safety constraints defined in~\eqref{eq:feas} and~\eqref{eq:safe} can be enforced through linear relationships in the polynomial coefficients.

A single polynomial is generally not sufficient to represent an entire trajectory. To increase flexibility, the trajectory is divided into consecutive segments, each parameterized by an individual polynomial. A general $M$-segment polynomial trajectory $p : [0, \tau_M] \mapsto \mathbb{R}^d$, with segment durations
\(T_1,\ldots,T_M\) and $\tau_m = \sum_{j=1}^{m} T_j$, is defined as
\begin{equation}
    p(t) =
\begin{cases}
\sum_{i=0}^{n} c_{1}^{i}\, \beta_{n}^{i}\!\left(s_1(t)\right), & t \in [0,\tau_1], \\
\sum_{i=0}^{n} c_{2}^{i}\, \beta_{n}^{i}\!\left(s_2(t)\right), & t \in [\tau_1, \tau_2], \\
\quad\quad\vdots & \quad\vdots \\
\sum_{i=0}^{n} c_{M}^{i}\, \beta_{n}^{i}\!\left(s_M(t)\right), & t \in [\tau_{M-1}, \tau_{M}].
\end{cases}
\end{equation}

where $\beta_n^i(s) = \binom{n}{i}\, s^i (1 - s)^{n-i}$ denotes the Bernstein basis of order $n$ and $t$ is scaled by the duration $T_m$ to yield a standard B\'ezier curve with $s_m(t) = (t - \tau_{m-1})/T_m \in [0,1]$.  The $i^\text{th}$ control point of the $m^\text{th}$ segment of the B\'ezier curve is parameterized by $c^i_m \in \mathbb{R}^d$.
\paragraph{Derivatives}
For a B\'ezier segment $p_m$, its derivative can be expressed by a linear combination of corresponding lower-order control points:
\begin{align} \label{eq:derivative}
   \dot p_m(t) =
\frac{n}{T_m} \sum_{i=0}^{n-1}
\left(c_m^{i+1} - c_m^i\right)
\beta_{n-1}^i(s_m(t)),
\end{align}
for $t \in [\tau_{m-1}, \tau_m]$,
which itself represents a B\'ezier curve with $n$ control points ${c_m^i}' = n / T_m \left(c_m^{i+1} - c_m^i\right)$ and basis $\beta_{n-1}^i(s)$. Thus, recursive application of this formula determines the control points of the higher-order derivatives.
\paragraph{Continuity and Boundary Constraints}
Continuous differentiability of the trajectory is enforced by imposing continuity constraints on the derivatives at segment boundaries:
\begin{align}
    p_m^{(j)}(\tau_{m+1}) = p_{m+1}^{(j)}(\tau_{m+1}),
    \label{eq:polycontinuity}
\end{align}
for all $j \in \{0, \dots, n_c\}$ and $m \in \{1,\dots,M-1\}$,
which constrains control points of consecutive segments through linear relationships. For the $m^\text{th}$ and $(m+1)^\text{th}$ segment, we present the expression for derivatives up to $2^\text{nd}$ order:
\begin{align} \label{eq:continuity_}
    &c_m^n = c_{m+1}^0, \nonumber\\
    &(c_m^n - c_m^{n-1}) / T_m = (c_{m+1}^1 - c_{m+1}^{0}) / T_{m+1}, \\
    &(c_m^n - 2 c_m^{n-1} + c_m^{n-2}) / T_m^2 = (c_{m+1}^2 - 2c_{m+1}^1 + c_{m+1}^{0}) / T_{m+1}^2 \nonumber
\end{align}
The boundary constraints can be expressed in an equivalent way, replacing one side of the equation with the fixed boundary derivatives.

\paragraph{Safety Constraints}
B\'ezier curves are widely used in constrained trajectory optimization due to their \textit{convex hull property}, which
guarantees that the curve lies entirely within the convex hull
of its control points. Consequently, constraining the control
points of each segment $\{c_m^i\}_{i=0}^n$ to lie within a convex set $k$ ensures that the entire trajectory segment remains inside that set:
\begin{align}
    \{c_m^i\}_{i=0}^n \in \mathcal{Q}_k \quad \rightarrow \quad p_m(t) \in \mathcal{Q}_k.
\end{align}
Given our safety constraint $p_m(t) \in \bigcup_{k \in \mathcal{K}} \mathcal{Q}_k$ defined over a union of convex sets, we will keep the assignment of segments to sets flexible which is discussed in the next section.

\paragraph{Dynamic Feasibility Constraints}
The convex hull property can be used equivalently to ensure boundedness of the trajectory derivatives:
\begin{align} \label{eq:feasibilityconstraint}
    \underline{b}_j \leq \{(c_m^i)^{(j)}\}_{i=0}^{n-j} \leq \overline{b}_j \quad \rightarrow \quad p_m^{(j)}(t) \in \mathcal{B}^j,
\end{align}
for all $j \in \{0,\dots,n\}$,
effectively constraining each derivative control point to remain in the bounded set $\mathcal{B}^j := \{c^{(j)} \in \mathbb{R}^d \,|\,  \underline{b}_j \leq c^{(j)} \leq \overline{b}_j\}$.
\paragraph{Smoothness Objective}
The integral of the squared $n$-th derivative over a segment can be expressed as a quadratic form in the polynomial coefficients:
\begin{align} \label{eq:objective}
    \int_0^{T_m} \left[p_m^{(n)}(t)\right]^2 dt \quad = \quad c_m^\top Q(T_m) c_m. 
\end{align}
with $c_m = [c_m^0, c_m^1, \dots, c_m^n]^\top$.

\paragraph{Time Allocation}
For a given time-allocation $\mathbf{T} = [T_1, T_2, \dots, T_m]$ both the continuity and boundary constraints \eqref{eq:continuity_} depend linearly on the control points $c = [c_0, c_1, \dots, c_M]^\top$ and can be written in the compact matrix form as $\mathbf{A}_{\text{cont}} c = 0, \: \mathbf{A}_{\text{bound}} c = q_0$, respectively. Likewise, the derivative inequality constraints in \eqref{eq:feasibilityconstraint} can be written as $\mathbf{A}_{\text{ineq}} c \in \mathcal{B}$. The quadratic objective \eqref{eq:objective} can be rewritten in block-diagonal structure $c^\top \mathbf{Q} \,c$, overall yielding a convex quadratic program (QP), except for the non-convex safety constraints (see \eqref{eq:problem_statement2}).

\subsection{Spatio-temporal Allocation Graph}
In the previous section, we introduced the piecewise trajectory but did not address the non-convex \textit{union of convex set} constraint. In the following, we introduce a mixed-integer (MI) formulation for this constraint and use it to derive the concept of a spatio-temporal allocation graph. This graph structure is one of the main results, ensuring the success and efficiency of our solver.
\begin{figure}[t]
    \centering
    \includegraphics[width=0.78\linewidth]{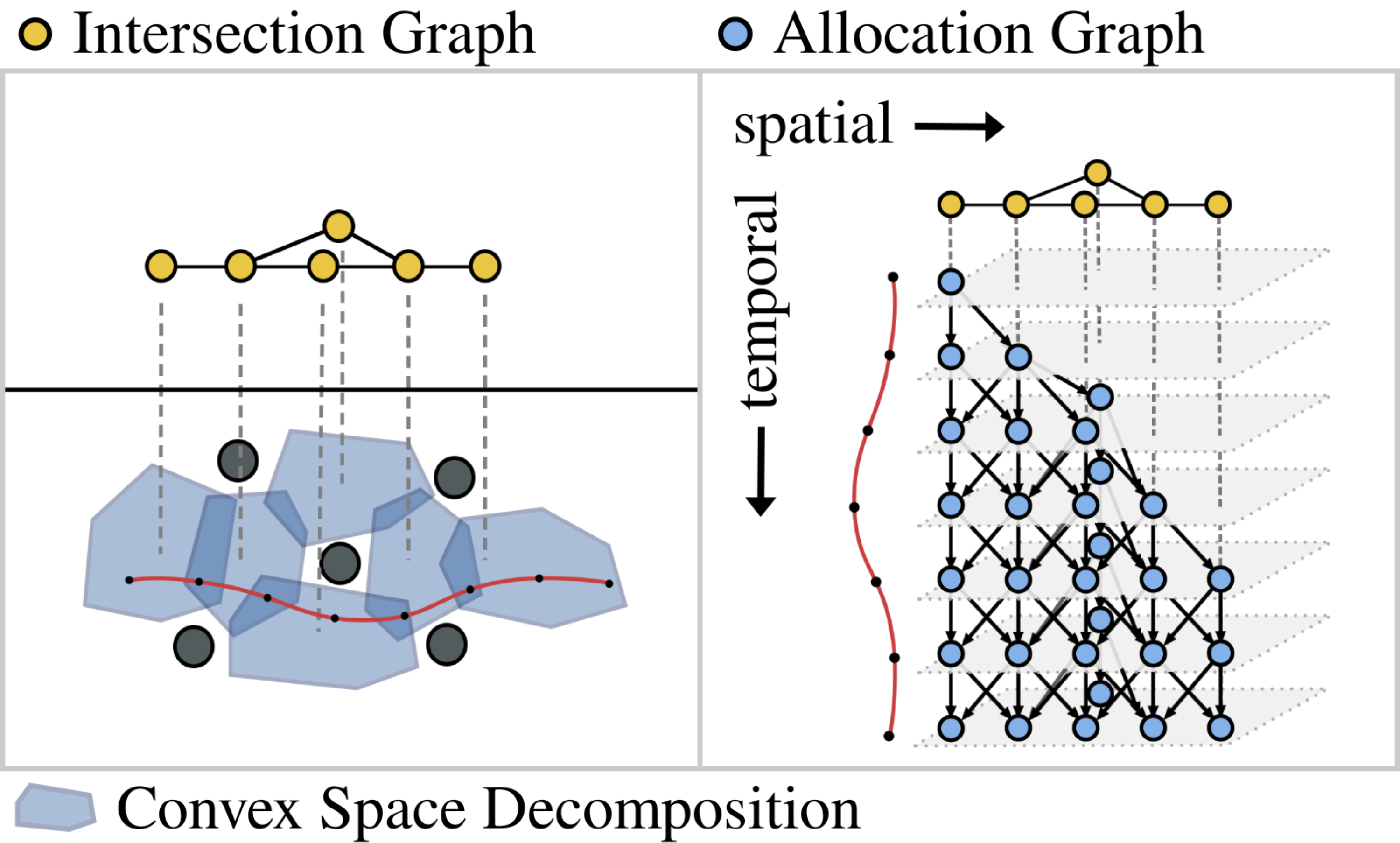}
    \caption{The allocation graph assigns segments (red) to convex sets (blue) and is constructed from the convex space decomposition and its corresponding intersection graph.}
    \label{fig:allocationgraph}
    \vspace{-2mm}
\end{figure}

\subsubsection{Mixed-Integer Formulation} \label{subsec:mi}
Let $\mathcal{M} := \{1,\dots, M\}$ index the polynomial segments.
We encode the assignment of each polynomial segment $m \in \mathcal{M}$ to a safe convex set $\mathcal{Q}_k$ using binary integer variables $b_{mk} \in \{0,1\}$:
\begin{subequations}
\begin{align}
    &b_{mk} \quad \Longrightarrow \quad \{c_m^i\}_{i=0}^n \subset \mathcal{Q}_k  \label{eq:integer} \\
    &\sum_{k=0}^{K} b_{mk} = 1 \label{eq:linearinteger}
\end{align}
\end{subequations}
where the linear constraint (\ref{eq:linearinteger}) ensures that segment $m$ is contained in at least one set $\mathcal{Q}_k$. Note that the implication in (\ref{eq:integer}) is one-directional and~\eqref{eq:linearinteger} only makes the assignment to one set explicit but allows a segment to be contained in multiple sets. \\
We further introduce binary variables $y_{m,(k,l)} \in \{0,1\}$ to indicate the continuous transition between sets through flow constraints:
\begin{subequations}
\begin{align}
    \sum_{l:(k,l) \in \mathcal{E}_\text{I}} y_{m,(k,l)} &= b_{m,k} \quad\forall m \in \mathcal{M} \setminus \{M\}, \forall k \in \mathcal{K} \label{eq:intflow1}\\
    \sum_{k:(k,l) \in \mathcal{E}_\text{I}} y_{m,(k,l)} &= b_{m+1,l} \: \forall m\in \mathcal{M} \setminus \{M\}, \forall l \in \mathcal{K}. \label{eq:intflow2}
\end{align}
\end{subequations}
Given the allocation of segment $m$ to set $k$ and segment $m+1$ to set $l$ expressed through $b_{m,k} = b_{m+1,l} = 1$, these constraints ensure that there is only one active flow variable $y_{m,(k,l)}$ connecting both sets. The flow constraints enforce consistency between assignments by restricting transitions to pairs of intersecting sets $(k,l) \in \mathcal{E}_\text{I}$ in the intersection graph $G_\text{I}$, which is defined in the next section.

\subsubsection{Intersection Graph $G_\text{I}$}
Let $\bigcup_{k \in \mathcal{K}} \mathcal{Q}_k$ denote a convex decomposition of the free space. We define the intersection graph $G_\text{I} = (\mathcal{V}_\text{I}, \mathcal{E}_\text{I})$ as follows:
\begin{align}
    \mathcal{V}_\text{I} &:= \mathcal{K} \\
    \mathcal{E}_\text{I} &:= \{(k,l) \in \mathcal{K} \times \mathcal{K} \,|\, \mathcal{Q}_k \cap \mathcal{Q}_l \neq \emptyset \}
\end{align}
where each convex set $\mathcal{Q}_k$ is associated with a vertex $k$ and edges exist between intersecting sets. This is depicted in Fig.~\ref{fig:allocationgraph}. The intersection graph encodes the adjacency structure of the sets $\{\mathcal{Q}_k\}_{k \in \mathcal{K}}$ and thus captures the connectivity of the free space $\mathcal{S}$. In particular, paths in $G_\text{I}$ correspond to sequences of overlapping convex regions that permit continuous transitions within $\mathcal{S}$.

\subsubsection{Allocation Graph $G_\text{A}$}\label{subsec:alloc_graph}
In the following, we show that the previously introduced mixed-integer formulation can be reformulated into a concise graph structure which we denote as allocation graph $G_\text{A} = (\mathcal{V}_\text{A}, \mathcal{E}_\text{A})$. We let each variable $b_{mk}$ represent a vertex $(m,k)$ in $\mathcal{V}_\text{A} := \{(m,k) \,|\, m \in \mathcal{M},\, k \in \mathcal{K} \}$ and each flow variable $y_{m,(k,l)}$ an edge in
$\mathcal{E}_\text{A} := \{\left((m,k), (m+1,l)\right) \, |\, (k,l) \in \mathcal{E}_\text{I} \}$. A visual example is given in Fig. \ref{fig:allocationgraph}.

The constraint in \eqref{eq:linearinteger} ensures only one set is selected for each segment $m$ which results in a layered structure of the graph. The mixed-integer flow conservation constraints (\ref{eq:intflow1}, \ref{eq:intflow2}) enforce that exactly one incoming and one outgoing edge is selected at each layer, thereby ensuring connectivity across consecutive layers through a single path.
The strict temporal ordering of the trajectory segments $0 < \tau_1 < \dots < \tau_M$ induces a natural orientation of the edges from layer $m$ to layer $m+1$. Thus, the resulting allocation graph is a directed acyclic graph (DAG). We will exploit this structure for efficient graph search in our solver.

We remark that each path $\mathcal{P} \in G_A$ in the graph corresponds to a feasible assignment. In particular, for a given path the safe set becomes convex and problem (\ref{eq:problem_statement2}) reduces to a convex QP.

\subsection{Problem Description}

Finally, combining the polynomial parameterization introduced in Section~\ref{subsec:polynomials} with the mixed-integer formulation through the allocation graph introduced in Section~\ref{subsec:alloc_graph} yields the following optimization problem:
\begin{subequations}\label{eq:problem_statement2}
\begin{align}
    \min_{c, \:\mathcal{P} \in G_\text{A}} \quad \frac{1}{2} c^\top& \mathbf{Q} \,c \label{eq:2obj}\\
    \text{s.t.} \quad
     \mathbf{A}_\text{eq} \, c &= \mathbf{b}, \label{eq:2eq} \\
     \mathbf{A}_\text{ineq} \, c &\in \mathcal{B}, \label{eq:2feas} \\
     c &\in \mathcal{S}_\mathcal{P} \label{eq:2safety},
\end{align}
\end{subequations}
with
\begin{align*}
    \mathcal{S}_\mathcal{P} &:= \{c\, |\, \{c_m^i\}_{i=0}^n \subset \mathcal{Q}_k, \: \forall (m,k) \in \mathcal{P}\}, \\
    \mathcal{B} &:= \{c \,|\, \underline{b} \leq c \leq \overline{b}\}, 
\end{align*}
and $\mathbf{A}_\text{eq} = [\mathbf{A}_\text{cont}^\top, \mathbf{A}_\text{bound}^\top]^\top$ and $\mathbf{b} = [0, b^\top]^\top$.

The problem only involves spatial parameters in the form of control points $c$. However, temporal flexibility is retained through the solver’s assignment of variable number of trajectory segments to each set in the spatio-temporal allocation graph $G_\text{A}$. Analogously, the graph structure encodes alternative feasible routes and retains the non-convex solution space.

\section{THE \acronym SOLVER}\label{sec:method}

\begin{figure*}[ht]
    \centering
    \includegraphics[width=1.0\textwidth]{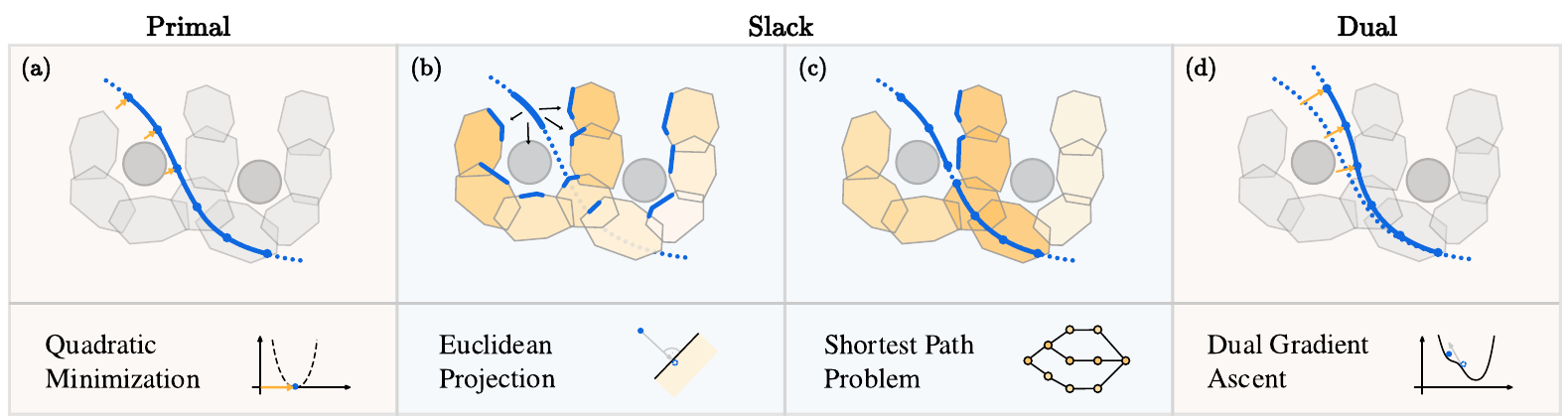}
    \caption{Effects of the ADMM updates on the trajectory. (a) The primal update yields a continuous trajectory. (b) The slack update projects each segment onto all convex sets. (c) A shortest-path search in the weighted allocation graph determines the segment allocation. (d) The dual update enforces consistency between primal and slack via a gradient ascent step.}
    \label{fig:admm}
    \vspace{-3mm}
    
\end{figure*}

The optimization problem in \eqref{eq:problem_statement2} involves a quadratic objective \eqref{eq:2obj} with affine equality \eqref{eq:2eq} and affine inequality constraints \eqref{eq:2feas} but, most importantly, a non-convex set constraint \eqref{eq:2safety} arising from the union of convex sets.

In our solver, we employ the Alternating Direction Method of Multipliers (ADMM) to address this problem in a systematic way. 
The key idea is to decompose the problem into subproblems that isolate the difficult set constraint from the smooth quadratic objective. This allows each update step to reduce to a problem class that is well understood and computationally efficient.

To enable this decomposition, we introduce auxiliary (slack) variables that decouple the quadratic objective from the set constraints. Specifically, we rewrite ~\eqref{eq:problem_statement2} in an equivalent consensus form
\begin{subequations} \label{eq:slackprob}
\begin{align}
    \min_{c} & & \frac{1}{2} c^\top \mathbf{Q} c &+ \mathbb{I}_{\mathbf{A}_\text{eq} c = \mathbf{b}}(c) + \mathbb{I}_{\mathcal{S}} (z_{\mathcal{S}}) + \mathbb{I}_{\mathcal{B}} (z_\mathcal{B}) \label{eq:slack_obj} \\
    \text{s.t.}
    && \mathbf{A}_\text{ineq} \, &c - z_\mathcal{B} = 0 \label{eq:zeq_c} \\
    && &c - z_{\mathcal{S}} = 0. \label{eq:zeq_nc}
\end{align}
\end{subequations}
with slack variables $z = (z_\mathcal{B}, z_{\mathcal{S}})$ and consensus constraints \eqref{eq:zeq_c}, \eqref{eq:zeq_nc} for both the convex and non-convex set, respectively. Here, $\mathbb{I}_\mathcal{X}$ denotes the indicator function of a set $\mathcal{X}$, so that $\mathbb{I}_\mathcal{X}(x) = 0$ for $x \in \mathcal{X}$ and $\mathbb{I}_\mathcal{X}(x) = \infty$ for $x \notin \mathcal{X}$. 

We proceed by formulating the Augmented Lagrangian (AL) of the equality-constrained problem~\eqref{eq:slackprob} from which the individual ADMM steps can be derived:
\begin{align}
    \mathcal{L}_A(c,z,\lambda) =&\: \frac{1}{2} c^T \mathbf{Q} c + \mathbb{I}_{\mathbf{A}_\text{eq} c = b} (c) + \mathbb{I}_{\mathcal{S}} (z_{\mathcal{S}}) + \mathbb{I}_{\mathcal{B}}(z_\mathcal{B}) \nonumber
    \\&+ \lambda_{\mathcal{S}}^T (c - z_{\mathcal{S}}) + \frac{\rho_{\mathcal{S}}}{2} ||c - z_{\mathcal{S}}||_2^2 \label{eq:lagrangian}
    \\&+ \lambda_\mathcal{B}^T (\mathbf{A}_\text{ineq} c - z_\mathcal{B}) + \frac{\rho_\mathcal{B}}{2} ||\mathbf{A}_\text{ineq} c - z_\mathcal{B}||_2^2 \nonumber
\end{align}
where $(\lambda_{\mathcal{S}}, \rho_{\mathcal{S}})$ and $(\lambda_\mathcal{B}, \rho_\mathcal{B})$ correspond to the dual variables and penalty parameters of the equality constraints ~\eqref{eq:zeq_nc} and ~\eqref{eq:zeq_c}, respectively.

The \textit{Method of Multipliers} performs primal-dual iterations on the Lagrangian by first minimizing w.r.t. the primal variables $c$ and $z$ before performing a dual-ascent step on the Lagrange multipliers $\lambda = (\lambda_\mathcal{B}, \lambda_\mathcal{S})$.
If we alternate minimization over $c$ and $z$, rather than simultaneously minimizing over both, we arrive at the three-step ADMM iteration:
\begin{subequations}
\begin{align}
    \text{primal update :} && c^+ &= \underset{c}{\arg\min} \; \mathcal{L}_A(c,z,\lambda), \label{eq:primalupdate} \\
    \text{slack update :} && z^+ &= \underset{z}{\arg\min} \; \mathcal{L}_A(c^+,z,\lambda), \label{eq:slackupdate}\\
    \text{dual update :} && \lambda^+ &= \lambda + \rho (c^+ - z^+). \label{eq:dualupdate}
\end{align}
\end{subequations}
These steps can be iterated until a desired convergence tolerance is achieved.

In the following, we detail each subproblem and provide its corresponding solution.

\subsubsection{\textbf{Primal} -- Equality-constrained QP}
The primal update solves an equality-constrained QP of the following form:
\begin{align} \label{eq:linearsys}
    \min_{c}\quad \frac{1}{2} c^\top \mathbf{\tilde{Q}} c + \tilde{q}^\top c \quad \text{s.t.} \quad \mathbf{A}_\text{eq}\, c = \mathbf{b}
\end{align}
with
\begin{align*}
    \mathbf{\tilde{Q}} &= \mathbf{Q} + \rho_{\mathcal{S}} \mathbf{I} + \rho_\mathcal{B} \mathbf{A}_\text{ineq}^\top \mathbf{A}_\text{ineq}  \nonumber \\
    \tilde{q} &= \lambda_{\mathcal{S}} - \rho_{\mathcal{S}} z_{\mathcal{S}} + \mathbf{A}_\text{ineq}^\top (\lambda_\mathcal{B} - \rho_\mathcal{B} z_\mathcal{B}) \nonumber.
\end{align*}
whose solution can be obtained by solving the linear system arising from the optimality conditions $\mathbf{A}_\text{eq}\, c^+ = \mathbf{b}$, $\mathbf{\tilde{Q}} c^+ + \tilde{q} + \mathbf{A}_\text{eq}^\top \lambda_\text{eq}^+ = 0$. Since both $\mathbf{\tilde{Q}}$ and $\mathbf{A}_\text{eq}$ remain unchanged between iterations, the factorization can be performed offline and online evaluation of $c^+$ only involves sparse matrix multiplications.

\subsubsection{\textbf{Slack} -- Projection \& Shortest-Path Problem}
The slack update \eqref{eq:slackupdate} takes the general form of an Euclidean projection problem
\begin{align}    
    \min_{z} ||\bar{z} - z||_2^2 \quad \text{s.t.} \quad z \in \mathcal{B} \times \mathcal{S}
\end{align}
where $\bar{z}$ denotes the projection point, obtained from the primal solution $c^+$ via an affine mapping, and $\mathcal{B} \times \mathcal{S}$ represents the feasible set. Intuitively, this step involves a projection of the primal variables onto the feasible set as illustrated in Fig. \ref{fig:admm}b. Since terms for $z_\mathcal{B}$ and $z_{\mathcal{S}}$ are fully separable, it is split into individual problems for each set.
\paragraph{Bounded Set ($\mathcal{B}$)}
For the convex set $\mathcal{B}$ this follows the standard ADMM formulation and the solution can be obtained in closed form through a simple linear projection onto the bounded set:
\begin{align}
    z_\mathcal{B}^+ = \Pi_\mathcal{B}(\bar{z}_\mathcal{B}) \label{eq:proj}
\end{align}   
with $\bar{z}_\mathcal{B} = \mathbf{A}_\text{ineq} c + \frac{1}{\rho_\mathcal{B}} \lambda_\mathcal{B}$. Evaluating $\Pi(z)$ can be performed efficiently by an element-wise min-max operation $\Pi_i(z_i) = \min\{\max\{z_i, \underline{b}\}, \overline{b}\}$.

\paragraph{Safety Set ($\mathcal{S}$)}
For the variables $z_{\mathcal{S}}$, the update involves a projection onto a non-convex set which may not be unique. In our setting, we have encoded all combinations of feasible segment-set projections in the spatio-temporal allocation graph $G_A$ that we introduced in (\ref{subsec:alloc_graph}). In accordance with \ref{eq:slackupdate}, we seek the minimizer of
\begin{subequations} \label{eq:projS}
\begin{align}
    \min_{z} &&& ||\bar{z} - z||_2^2\\
     \text{s.t.} &&& \{z^i_m\}_{i \in \{0, n\}} \subset \mathcal{Q}_k, \quad (m,k) \in \mathcal{P}, \: \mathcal{P} \in G_\text{A} \label{eq:projS_con}
\end{align}
\end{subequations}
with $\bar{z} = c$ . We have omitted the subscript of $z_{\mathcal{S}}$ for clarity.
The separable structure of~\eqref{eq:projS_con} across segments-set assignments $(m, k)$ and control points~$i$ admits the decomposition into a sum of independent contributions:
\begin{align}    
    \min_{\mathcal{P} \in G_A} \sum_{(m,k) \in \mathcal{P}} \sum_{i=0}^n \min_{z_m^i \in \mathcal{Q}_k} ||\bar{z}_m^i - z_m^i||_2^2, \label{eq:sumproj}
\end{align}
which is equivalent to a \textit{Shortest-Path Problem} (SPP) involving the accumulated projection costs for each segment-set pair in a possible path $\mathcal{P}  = \{(m, k)\}_{m=1}^M$. In Fig. \ref{fig:admm}, we visualize the cost for both segment-set pairs in (b) and paths in (c), respectively.

Since $\mathcal{Q}_k$ is convex, the projection of each control point onto a set $k$ can be done in closed-form as in ~\eqref{eq:proj}:
\begin{align}
    {z_{mk}^i}^+ = \Pi_{\mathcal{Q}_k}(\bar{z}_m^i) \label{eq:projsetq}
\end{align}
from which the projection costs for each segment-set pair $(m,k)$ is computed as $l_{mk} = \sum_{i=0}^n ||\bar{z}_m^i - {z_{mk}^i}^+||_2^2$.

This allows us to simplify problem \eqref{eq:sumproj} as follows:
\begin{align}
   \min_{\mathcal{P} \in G_A} \sum\nolimits_{(m,k) \in \mathcal{P}} l_{mk}
\end{align}
yielding a SPP on the weighted allocation graph $G_A$ with vertex cost $l_{mk}$.

\paragraph{Shortest-Path Problem on DAG} \label{subsec:SSP}
As derived in \ref{subsec:alloc_graph}, the allocation graph $G_A$ is a Directed Acyclic Graph (DAG) with layers induced through segments $\mathcal{M}$. Thus, the Shortest-Path Problem can be solved through $M$ dynamic programming iterations in topological order. The recursion is initialized at the first layer by setting $V_{1k} = l_{1k}$ for $k\in \mathcal{K}_\text{start}$, and $V_{1k} = +\infty$ otherwise.
For each $m = 2, \dots, M:$
\begin{subequations}
\begin{align}
    V_{mk} &= l_{mk} + \min_{l: (l, k) \in \mathcal{E}_I} V_{m-1, l}, &\forall k \in \mathcal{K}, \\
    \pi_{mk} &= \arg \min _{l: (l, k) \in \mathcal{E}_I} V_{m-1, l}, &\forall k \in \mathcal{K},
\end{align}
\end{subequations}
where $V_{mk}$ is the value function and $\pi_{mk}$ the minimizing predecessor of each node. After the forward pass, the optimal path can be obtained by simple backtracking from the goal node by $k_{m-1} = \pi_{m,k_m}$.

The optimal solution $z_{\mathcal{S}}^+$ is obtained by selecting $z_{mk}^+ = \{{z^i_{mk}}^+\}_{i=0}^n$ from \eqref{eq:projsetq} based on the optimal path $\{(m,k)\}_ \mathcal{P}$.

\subsubsection{\textbf{Dual} - Gradient-ascent Step}
After updating both primal variables $c$ and $z$ that minimize the AL, the dual update \eqref{eq:dualupdate} performs a gradient-ascent step on the Lagrange multipliers with step-size $\rho$.
This update gradually enforces consensus between variables $c$ and $z$. Intuitively, it acts as a price update that shifts the primal solution towards feasibility as depicted in Fig. \ref{fig:admm}d.

\subsection{Overview}
The method iterates between primal, slack, and dual updates until a prescribed convergence tolerance is achieved.
A key practical advantage of this procedure is that each step is computationally lightweight. The primal update admits an efficient evaluation of the QP solution, while the slack update reduces to projection and graph search operations whose complexity scales linearly with the number of sets and trajectory segments, respectively.
The method requires only a single set of hyperparameters, the penalty parameters $\rho_{\mathcal{S}}$ and $\rho_\mathcal{B}$, which correspond to the step-size of the dual ascent step.

\subsection{Convergence and Completeness}
Unlike the convex case, there is no general guarantee that ADMM will converge to a globally optimal point for non-convex set constraints and it must be considered a local optimization method \cite{takapoui2020simple}. However, our method encodes all \textit{feasible}\footnote{feasibility here only refers to the non-convex constraint $\mathcal{S}$} combinations of the MIQP \eqref{eq:problem_statement2} in the allocation graph $G_A$ and searches for the locally optimal one in each iteration \ref{subsec:SSP}. Recall that each path $\mathcal{P} \in G_A$ corresponds to a convex QP based on our MI formulation in \ref{subsec:mi}. Thus, if the path $\mathcal{P}$ is frozen at any time, we recover the convex setting, and primal and dual residuals are guaranteed to converge to 0 in the limit \cite{stellato2020osqp}. In practice, we observe that this happens naturally after a few tens of iterations and does not have to be enforced explicitly.
The primal variables $c$ and $z$ will converge to stationary points under this condition, even if the bounded set $\mathcal{B}$ renders some QPs in $G_A$ infeasible. Infeasible solutions are naturally avoided in our algorithm by selecting the locally optimal path $\mathcal{P}$ in each iteration, but general completeness of the algorithm can only be guaranteed for the unbounded case.

For the special case, where derivatives of the boundary conditions (start and goal states) are zero, we can solve the unbounded problem and afterwards scale the total time to satisfy any bounds on the derivatives of the trajectory.

\section{EXPERIMENTS}\label{sec:experiments}
We demonstrate the effectiveness of ACTOR on a variety of numerical examples. 
Section \ref{subsec:qualitative} illustrates, through qualitative comparisons, the key advantages of ACTOR over existing state-of-the-art approaches. We then demonstrate the method on a navigation task in Section \ref{subsec:naivespace}.
Section \ref{subsec:scalability} examines scalability and computational complexity.
Finally, we highlight the advantages of our method in a challenging study on minimum-snap trajectory generation under higher-order feasibility constraints.

As representative baselines, we consider the following methods from the literature:
\begin{itemize}
    \item Optimization-based Collision Avoidance (\textbf{OBCA}) \cite{zhang2020optimization}, a direct NLP formulation that enforces collision avoidance through direct nonlinear obstacle constraints. The resulting problem is solved using IPOPT \cite{biegler2009large}.
    \item Safe corridor optimization (\textbf{SafeC}), as employed in several prior works \cite{richter2016polynomial, wang2022geometrically, von2024using, tordesillas2021faster, ren2025safety} which optimizes a spatio-temporal trajectory within a prescribed sequence of convex sets. Following \cite{von2024using}, we use SNOPT \cite{gill2005snopt} as the NLP solver.
    \item \textbf{A*}, used as path search heuristic both to initialize OBCA and to construct safe corridors for SafeC \cite{ren2025safety}.
    \item Graph of Convex Sets (\textbf{GCS}) \cite{marcucci2023motion}, used as a convex surrogate method for safe corridor selection.
\end{itemize}
We use the same polynomial parameterization and objective for all methods and rely on optimized C/C++ implementations. Only our method is implemented in Python/\textit{NumPy}.

\subsection{Non-convex Environments} \label{subsec:qualitative}
\begin{figure}[t]
    \centering
    \begin{subfigure}{0.22\textwidth}
        \centering
        \includegraphics[width=\linewidth]{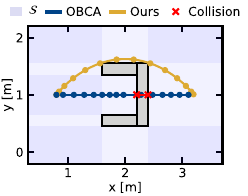}
        \caption{Local Minimum.}
        \label{fig:culdesac}
    \end{subfigure}
    \begin{subfigure}{0.25\textwidth}
        \centering
        \includegraphics[width=\linewidth]{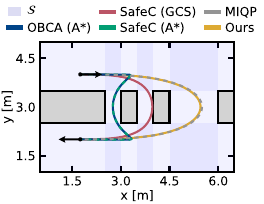}
        \caption{Multiple Paths.}
        \label{fig:detour}
    \end{subfigure}
    \caption{Illustrative examples of challenging non-convex environments highlighting the benefits of our method.}
    \label{fig:both}
\end{figure}
Non-convex space constraints pose a fundamental challenge in trajectory optimization, as they induce infeasible local minima and multiple homotopy classes.
To demonstrate the conceptual benefits of our method, we consider two minimal illustrative scenarios that are specifically designed to expose characteristic weaknesses of existing trajectory optimization approaches.
The first scenario depicted in Fig. \ref{fig:culdesac} consists of a U-shaped obstacle positioned between start and goal.
Both methods are initialized with a straight-line that enters the cavity of the obstacle. Escaping this cavity requires a global deformation that temporarily increases cost and constraint violation. OBCA follows local descent directions and converges to an infeasible stationary point from which it cannot escape, failing to produce a feasible trajectory.
In ACTOR, each gradient step is coupled with the allocation graph, which encodes global connectivity of the free space and systematically guides the iterates toward a feasible homotopy class. Importantly, this behavior extends beyond minimal examples: in the highly non-convex maze shown in Fig. \ref{fig:quad}, ACTOR reliably converges to feasible trajectories despite the presence of numerous competing homotopy classes.

To address this limitation, many existing methods generate a feasible initialization or pre-select a convex safe corridor prior to optimization \cite{ren2025safety, von2024using}. While often effective, this strategy can yield suboptimal solutions by restricting the admissible solution space. 
We exemplify this in the second example in Fig. \ref{fig:detour}, which features multiple routes through a separating wall. In this scenario, we consider trajectories subject to dynamic initial and final state constraints and determine the optimal solution by solving the corresponding MIQP to global optimality.
Both OBCA and SafeC when initialized from the shortest-path based on A* yield suboptimal trajectories involving sharp turns and aggressive maneuvers. GCS selects a corridor based on convex approximations for the smooth nonlinear constraints and objective of the problem which results in a better homotopy choice and a smoother trajectory. However, the lack of tightness of these surrogates still affects the optimality of the solution. Only ACTOR converges to the globally optimal trajectory by jointly taking spatial and dynamic constraints into account. 

\subsection{Naive Space Decomposition} \label{subsec:naivespace}
\begin{figure}[t]
        \centering
        \includegraphics[width=\linewidth]{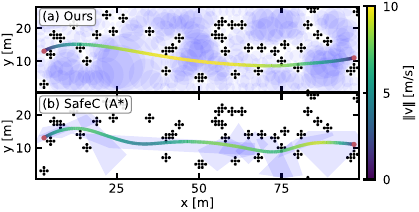}
        \caption{Continuous trajectories in a point-cloud environment.}
        \label{fig:naive}
    \label{fig:both}
\end{figure}
Having demonstrated the key qualitative properties of ACTOR in the previous section, we now examine how these properties translate to practical navigation tasks. 
We adopt a point-cloud representation of the obstacles to highlight a key practical benefit of the proposed union-of-convex-sets formulation: free space can be decomposed in a simple and naive manner. Unlike existing methods, whose performance depends critically on carefully constructed free-space decompositions around a prescribed initial path, ACTOR can operate effectively on such naive decompositions. As shown in Fig. \ref{fig:naive}, simply inflating circular sets on a grid and leveraging the proposed graph formulation enables a much larger admissible solution space, allowing ACTOR to recover a substantially smoother trajectory than a corridor-based approach (SafeC), whose solution remains restricted to the pre-selected safe corridor.
This ability of our method to operate on naive decompositions eliminates the need to pose free-space decomposition as an additional optimization problem, substantially simplifying the planning pipeline.

\subsection{Scalability} \label{subsec:scalability}
\begin{figure}[b]
    \centering
    \begin{subfigure}[t]{0.2\textwidth}
        \centering
        \includegraphics[width=\linewidth]{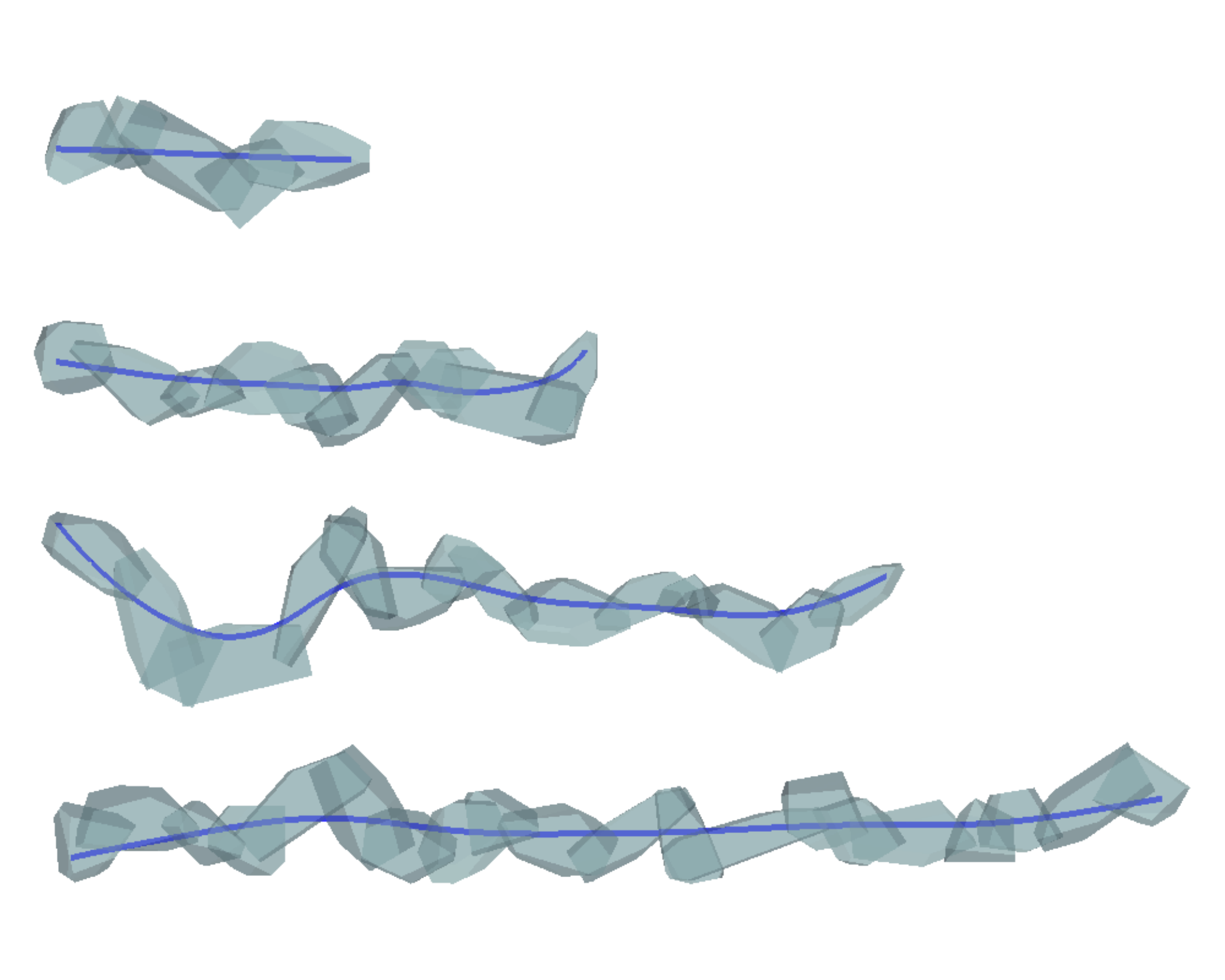}
    \end{subfigure}
    \begin{subfigure}[t]{0.22\textwidth}
        \centering
        \includegraphics[width=\linewidth]{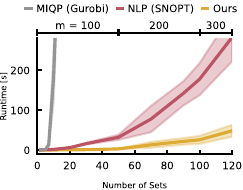}
    \end{subfigure}
    \caption{Left: Trajectories in sequences of sets. Right: Runtime as a function of the number of sets for different solvers.}
    \label{fig:runtime}
\end{figure}
While the previous subsection established the practical relevance of ACTOR, we now investigate its scalability as the size of the problem increases.
For a controlled comparison with existing safe-corridor methods, we consider problems defined over a single prescribed corridor and examine the resulting scaling behavior as the number of sets and trajectory segments increases.
Within this setting, we compare ACTOR against a commercial MIQP solver, solving the same MIQP formulation as ours to global optimality as in \cite{tordesillas2021faster}, and the NLP-based spatio-temporal SafeC formulation of \cite{von2024using}.
As shown in Fig. \ref{fig:runtime}, runtimes for both the MIQP and NLP solver grow combinatorially or exponentially due to the repeated solution of increasingly large QPs. In contrast, ACTOR is factorization-free and exhibits per-iteration complexity linear in the number of sets and trajectory segments, thereby scaling substantially better to large problem instances.

\subsection{Quadrotor Minimum-Snap Trajectory Planning} \label{subsec:quadmaze}
\begin{figure}[t]
    \centering
    \includegraphics[width=\linewidth]{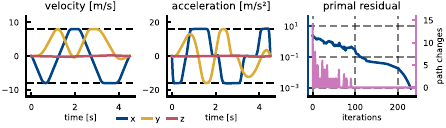}
    \caption{Bounded velocity and acceleration profile and convergence for the min-snap trajectory in Fig. \ref{fig:quad}.}
    \label{fig:convergence}
\end{figure}
Finally, we showcase the full strength of ACTOR on a challenging quadrotor minimum-snap planning problem that combines non-convex spatial constraints with higher-order dynamic feasibility constraints on velocity and acceleration.
As shown in Fig. \ref{fig:quad}, our method is able to find a smooth and feasible trajectory in such a highly constrained setting, where most paths in the allocation graph are rendered infeasible by the constraints on velocity and acceleration. In fact, we observe reliable convergence from a naive initial guess to a feasible optimum in these settings as depicted in Fig.~\ref{fig:convergence}. The number of discrete vertex changes in the allocation path stabilizes after 100 iterations, from which we retain the convex setting and rapid convergence to a pre-specified error tolerance.
This experiment consolidates the main advantages of the proposed method: ACTOR jointly optimizes over continuous spaces involving the higher-order dynamics and discrete spaces involving non-convex geometry, avoiding the restrictive decoupling inherent to existing pipeline-based approaches.

\section{CONCLUSION}\label{sec:conclusion}
In this work, we introduced ACTOR, a novel solver for continuous trajectory optimization in non-convex environments described by general unions of convex sets. 
Central to the method is a spatio-temporal graph which enables the joint treatment of smooth dynamics and discrete constraints within a single ADMM-based optimization scheme.
The resulting solver is robust to naive initialization, factorization-free, and exhibits linear scaling with respect to problem size. Extensive experimental results demonstrate its effectiveness, underscoring the potential of this framework as a principled and scalable foundation for motion planning and trajectory optimization in increasingly complex environments.

\addtolength{\textheight}{-12cm}   





\printbibliography{}

\end{document}